\documentclass[letterpaper, 10pt, conference]{ieeeconf}  % Comment this line out if you need a4paper

\IEEEoverridecommandlockouts                              % This command is only needed if
\usepackage{amsmath}
\usepackage{amssymb}
\usepackage{booktabs}
\usepackage{mathtools}

\usepackage[utf8]{inputenc} %% needs to go before biblatex and csquotes

\usepackage{calc}
\usepackage[font=small,skip=1pt]{caption}
\usepackage{circuitikz}
\usepackage{csquotes}
\usepackage{color}
\usepackage{colortbl}
\usepackage{xspace}

\let\labelindent\relax % to fix compilation with enumitem
\usepackage[inline]{enumitem}
\usepackage[T1]{fontenc}
\usepackage[nolist, withpage, smaller]{acronym}
\usepackage{graphicx} % Required for inserting images
\usepackage[unicode=true,
	hidelinks,
	pdfsubject={},
	pdfkeywords={},
	colorlinks=true,
	allcolors=blue,
	citecolor=blue
]{hyperref}

\usepackage{ifthen}

\usepackage{lipsum}

\usepackage{multirow}
\usepackage{multicol}

\usepackage{pgffor}
\usepackage{pgfgantt}
\usepackage{pgfplots, pgfplotstable}
\usepackage{rotating}
\usepackage{siunitx}
\usepackage{subcaption}
\usepackage{tabularx}
\usepackage{tikz}
\usepackage{tikzsymbols}
\usepackage{verbatim}
\usepackage{xargs}
\usepackage{xcolor}
\usepackage{xifthen}

\usepackage[capitalise]{cleveref}

\usetikzlibrary{arrows}
\usetikzlibrary{arrows.meta}
\usetikzlibrary{automata}

\usetikzlibrary{backgrounds}
\usetikzlibrary{bending}

\usetikzlibrary{calc}
\usetikzlibrary{calendar}
\usetikzlibrary{chains}
\usetikzlibrary{circuits.logic.CDH}
\usetikzlibrary{circuits.logic.US}

\usetikzlibrary{decorations}
\usetikzlibrary{decorations.footprints}
\usetikzlibrary{decorations.fractals}
\usetikzlibrary{decorations.markings}
\usetikzlibrary{decorations.pathmorphing} % noisy shapes
\usetikzlibrary{decorations.pathreplacing}
\usetikzlibrary{decorations.shapes}
\usetikzlibrary{decorations.text}

\usetikzlibrary{er}

\usetikzlibrary{fadings}
\usetikzlibrary{fit}
\usetikzlibrary{folding}

\usetikzlibrary{matrix}
\usetikzlibrary{mindmap}

\usetikzlibrary{patterns}
\usetikzlibrary{petri}
\usetikzlibrary{pgfplots.colorbrewer} % LATEX and plain TEX
\usetikzlibrary{plotmarks}
\usetikzlibrary{positioning}

\usetikzlibrary{scopes}
\usetikzlibrary{shadows}
\usetikzlibrary{shapes}
\usetikzlibrary{shapes.arrows}
\usetikzlibrary{shapes.callouts}
\usetikzlibrary{shapes.gates.logic.US}
\usetikzlibrary{shapes.gates.logic.IEC}
\usetikzlibrary{shapes.geometric}
\usetikzlibrary{shapes.misc}
\usetikzlibrary{shapes.multipart}
\usetikzlibrary{shapes.symbols}

\usetikzlibrary{through}
\usetikzlibrary{tikzmark}
\usetikzlibrary{topaths}
\usetikzlibrary{trees}

\usepgfplotslibrary{colormaps}
\usepgfplotslibrary{colorbrewer} % LATEX and plain TEX
\usepgfplotslibrary{dateplot}
\usepgfplotslibrary{fillbetween}
\usepgfplotslibrary{patchplots}
\usepgfplotslibrary{groupplots}
\usepgfplotslibrary{statistics}

\pgfplotsset{compat=1.18}

\usepackage[colorinlistoftodos]{todonotes}

\newif\ifdraft
	\definecolor{ColorHP}{RGB}{100,209,255}
	\definecolor{ColorNF}{RGB}{190,174,255}
	\definecolor{ColorGJ}{RGB}{204, 153, 0}

\newcommand{\gj}[1]{\ifdraft{\todo[inline, color=ColorGJ!20, bordercolor=ColorGJ!0]{\scriptsize\textbf{Georg:} #1}}\fi}

\newcommand{\cmpr}[1]{cf.~\cref{#1}}

\newcommand{\mycite}[1]{\cite{#1}}%\ifdraft{\textcolor{red}{\cite{#1}}}\else{\cite{#1}}\fi}

\newcommand{\plant}{\mathcal{P}}%

\newcommand{\decision}{\mathcal{D}}

\newcommand{\shift}{1cm}

\colorlet{ColorComplex}{white}
\colorlet{ColorDecision}{black!60}
\colorlet{ColorSimplex}{white}
\colorlet{ColorVoter}{ColorDecision}

\tikzset{arcnode/.style={rectangle, draw=black, rounded corners=0.125cm, minimum height=0.33*\shift, align=center, fill=white}}
\tikzset{switchhighlightnode/.style={draw=red!80!black!20!white, fill=red!80!black!20!white, line width=0.25mm, minimum width=8.5*\shift, minimum height=3.75*\shift, rounded corners}}
\tikzset{flowedge/.style={-latex, black, line width=0.25mm}}

\pgfdeclarelayer{farback}
\pgfsetlayers{farback,background,main}

\NewDocumentCommand{\Sensor}{ O{0} O{1.0} O{Sensor} O{black} m}{
	\begin{scope}[scale=#2]
		\node[anchor=south, draw, fill=#4!20, minimum width=0.2cm, minimum height=0.2cm, inner sep=0pt, scale=#2, rotate=#1] (SBox) at (#5) {};
		\node[minimum width=0.2cm, minimum height=0.4cm, inner sep=0pt, scale=#2, rotate=#1] (#3) at (SBox.north) {};

		\coordinate (A) at ($(SBox.north)+({-0.1*cos(-#1)+0.2*sin(-#1)},{0.1*sin(-#1)+0.2*cos(-#1)})$);
		\coordinate (B) at ($(SBox.north)+({0.1*cos(-#1)+0.2*sin(-#1)},{-0.1*sin(-#1)+0.2*cos(-#1)})$);
		\draw[fill=#4!40] (SBox.north) -- (A) -- (B) -- cycle;

	\end{scope}
}

\usepackage{ifthen}
\newboolean{blinded}
\setboolean{blinded}{false}

\title{\LARGE \bf
Towards Safe Path Tracking Using the Simplex Architecture
}

\ifthenelse{\boolean{blinded}}{
  \author{Blinded% <-this % stops a space
  \thanks{Blinded}%
  }
}{
  \author{Georg Jäger$^{1}$, Nils-Jonathan Friedrich$^{1}$, Hauke Petersen$^{2}$, Benjamin Noack$^{2}$% <-this % stops a space
  \thanks{This work was partially funded by the German Federal Ministry for Digital and Transport within the mFUND program (grant no. 19FS2025A - Project Ready for Smart City Robots).}% <-this % stops a space
  \thanks{$^{1}$Georg Jäger and Nils-Jonathan Friedrich are with the Faculty of Mathematics and Computer Science,
          TU Bergakademie Freiberg, 09599 Freiberg, Germany
          {\tt\small \{georg.jaeger, nils-jonathan.friedrich\}@informatik.tu-freiberg.de}}%
  \thanks{$^{2}$Hauke Petersen and Benjamin Noack are with the Faculty of Computer Science, Otto-von-Guericke University Magdeburg, 39106 Magdeburg, Germany
          {\tt\small \{hauke.petersen, benjamin.noack\}@ovgu.de}}%
  }
}

\begin{document}

\maketitle
\thispagestyle{empty}
\pagestyle{empty}

\begin{abstract}
	Robot navigation in complex environments necessitates controllers that prioritize safety while remaining performant and adaptable.
Traditional controllers like Regulated Pure Pursuit, Dynamic Window Approach, and Model-Predictive Path Integral, while reliable, struggle to adapt to dynamic conditions.
Reinforcement Learning offers adaptability but state-wise safety guarantees remain challenging and often absent in practice.
To address this, we propose a path tracking controller leveraging the Simplex architecture.
It combines a Reinforcement Learning controller for adaptiveness and performance with a high-assurance controller providing safety and stability.
Our main goal is to provide a safe testbed for the design and evaluation of path-planning algorithms, including machine-learning-based planners.
Our contribution is twofold.
We firstly discuss general stability and safety considerations for designing controllers using the Simplex architecture.
Secondly, we present a Simplex-based path tracking controller.
Our simulation results, supported by preliminary in-field tests, demonstrate the controller's effectiveness in maintaining safety while achieving comparable performance to state-of-the-art methods.

\end{abstract}

% \begin{IEEEkeywords}
% 	component, formatting, style, styling, insert
% \end{IEEEkeywords}

%%%%
%% Introduction
%%%%
\section{Introduction}
\label{sec:introduction}

\gj{Ideas to strengthen the paper: Motivate adaptivity due to versatile driving scenarios (complex trajectories) using proposed path tracking controller. PP was extended to be adaptive in multiple ways (fuzzy, clothoid fitting, RPP), MPPI employs optimization to adapt to the environment but is hard to provide safety guarantees due to complexity, adaptive PID controllers, adaptive Stanley controller, etc. The central motivation for these adaptations is the need for adaptivity in complex environments. Can we find examples visualizing this problem? Not only for (R)PP? Then, transition to intelligent controllers promising adaptivity (e.g. fuzzy logic, neural networks, RL-based), but lacking safety guarantees. Then, state our contribution as a way to construct a simplex architecture for safe employment of performance-oriented controllers (e.g. RL) with the use case of path tracking. It will allow us to execute the RL controller within the safety envelope of the high-assurance controller (e.g. RPP), which means, we are still limited by the performance of the high-assurance controller in the worst case, but can achieve better performance in the best case.}

\begin{figure}[!h]
	\centering
	\includegraphics[width=0.75\linewidth]{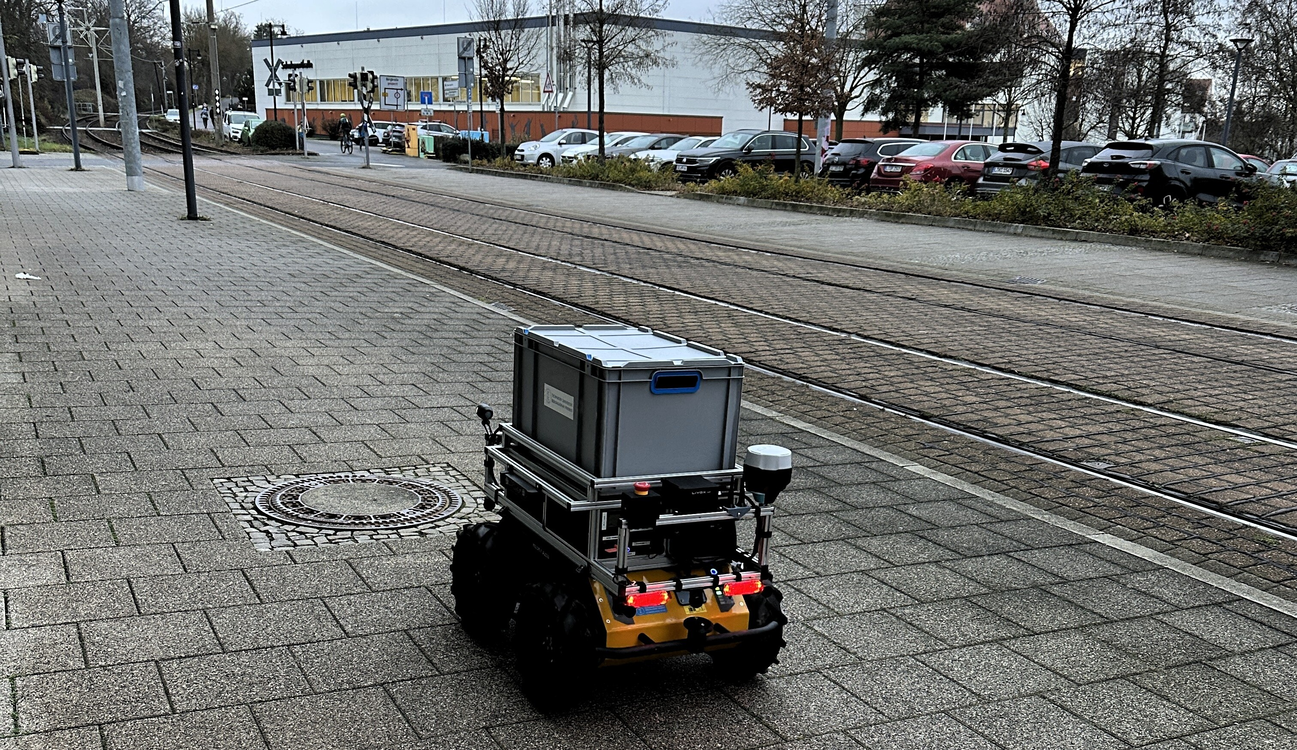}
	\caption{Autonomous mobile robot demonstrating parcel delivery on a sidewalk of 2m width.}
	\label{fig:robot_sidewalk}
\end{figure}
Autonomous mobile robots will revolutionize applications, e.g., in parcel delivery~\cmpr{fig:robot_sidewalk}, agriculture, healthcare, etc. \mycite{cognominal2021evolving}.
A central prerequisite for that is safe and effective navigation, which requires robots to follow a (pre-)planned path.
This task is typically addressed by path tracking algorithms which are constantly challenged by complex trajectories.
The \ac{pp} algorithm~\mycite{ollero95stability} formulates a geometric approach to follow a path by steering towards a lookahead point on the path.
Although proven to be stable~\mycite{ollero95stability}, it's assumption of constant velocity and curvature raised the question for adaptivity, which led to the \ac{app} and \ac{rpp} algorithms~\mycite{macenski2023regulated}.

In a similar vein, PID controllers have been extended to track velocity and provide steering control \mycite{mai2021combined} using fuzzy logic.
The predictive \textit{Stanley} controller has been extended using fuzzy logic to provide adaptivity regarding its prediction horizon \mycite{abdelmoniem2023fuzzy}.
\ac{mppi} \mycite{williams2016aggressive} employs an optimization-based approach to adaptively compute control commands while considering the robot's dynamics and constraints.
However, static, risk-neutral trajectory sampling limits its adaptivity and is extended to use uncertainty for efficient trajectory sampling in \mycite{mohamed2025toward}.
\mycite{lin2019path}, similarly, provides adaptivity to the underlying model of the system dynamics by firstly adapting the tire cornering stiffness and the road friction coefficient, and secondly, using \ac{mpc} to compute control commands.
\ac{rl} offers an alternative and has proven its performance and adaptability for path tracking~\mycite{cao2024path,chai2024design,wang2020trajectory} as well as in other complex, dynamic application scenarios~\mycite{singh2022reinforcement}.
However, the \textit{black-box} nature of \ac{ann} commonly used in (deep) \ac{rl} contradicts the a central motivation and need for adaptivity in path tracking: safety and stability guarantees.
Although this challenge is most pronounced in for \ac{ann}-based controllers, it also applies to other adaptive controllers, where increasing complexity increases the difficulty to provide guarantees (e.g. consider optimization-based model-predictive controllers).

Hybrid architectures, such as the Simplex architecture \mycite{sha2001using}, aim to bridge this gap by enforcing a safety envelope around an uncertified \ac{rl} controller, with the safety guarantees of traditional controllers.
The \ac{rl} controller operates under normal conditions, while a high-assurance fallback controller (e.g., \ac{dwb}, \ac{mppi}, or \ac{rpp}) ensures safety when needed.
In this way, safety is guaranteed, independently of the high‑performance controller’s degree of optimality.

Despite their potential, hybrid architectures seem to be underexplored in path tracking and autonomous navigation, most notably in \ac{ros2}.
To address this gap, we make two contributions.
Firstly, we discuss design considerations for ensuring safety and stability in controllers using the Simplex architecture (\cmpr{sec:related_work}).
Secondly, we present a Simplex-based path tracking controller (\cmpr{sec:concept}) as a safety-wrapper for deploying complex, adaptive controllers.
We evaluate the controller using \ac{rl} as the high-performance controller and \ac{rpp} as the high-assurance controller in simulations (\cmpr{sec:evaluation}), demonstrating its effectiveness in maintaining safety while achieving comparable performance to state-of-the-art methods, as well as in a real-world deployment.
In the next section, however, we start by discussing the limitations of existing path tracking controllers in \ac{ros2} and approaches for introducing adaptiveness through \ac{rl} (\cmpr{sec:state_of_the_art}).

%%%%
%% State of the Art
%%%%
\section{State of the Art on Path Tracking Controllers in ROS2}
\label{sec:state_of_the_art}

\ac{ros2} path tracking controllers can be classified as geometric, reactive, or model-predictive \mycite{macenski2023from}.
This section reviews \ac{rpp}, \ac{dwb}, and \ac{mppi} as commonly-used examples of these categories.
We focus on their applicability to open, dynamic environments, and discuss their limitations in adaptability, performance, and applicability.
As \ac{rl} addresses these properties, we also discuss its potential in path tracking (\cmpr{subsec:rl}) before concluding the section with a discussion (\cmpr{subsec:discussion}).

\subsection{Regulated Pure Pursuit (RPP)}
\label{subsec:rpp}

\ac{rpp}~\mycite{macenski2023regulated} is a geometric controller derived from \ac{pp} algorithm that identifies a \textit{lookahead point} at distance $L$ ahead on the path to calculate curvature and angular velocity.
This basic algorithm is asymptotically stable for specific lookahead distances~\mycite{ollero95stability}.
It's proof extends to the extensions introduced by the \ac{rpp} version, which adjust the lookahead distance based on path curvature and adjust the velocity based on obstacle proximity to enhance performance.

% \ac{rpp}~\mycite{macenski2023regulated} is a geometric controller based on the pure pursuit algorithm.
% It determines a so-called \textit{lookahead point} on the planned path at a specified lookahead distance $L$ in front of the robot.
% Then, it calculates the curvature and derives the angular velocity from it.
% This basic pure pursuit algorithm has been shown to be asymptotically stable for a range of lookahead distances~\mycite{ollero95stability} that depends on the robot's velocity.
% Thus, to optimize performance, the \ac{rpp} variant introduces a curvature-dependent lookahead distance~\mycite{paden2016survey}.
% It is further extended in \mycite{macenski2023regulated} to also consider the robot's distance to obstacles for adaptively adjusting the target velocity.
% As these adaptations are based on geometric considerations, \mycite{ollero95stability} still holds for \ac{rpp}, ensuring stability.

\subsection{Dynamic Window Approach Version B (DWB)}
\label{subsec:dwb}

\ac{dwb} \mycite{fox1997dynamic} is a reactive controller that samples linear and angular velocities within a \textit{dynamic window} of feasible motions to generate circular, collision-free trajectories.
Each is scored by goal proximity, obstacle distance, and current velocity, with the lowest-cost control applied.
Although this integrates feasibility and safety, its fixed cost function requires careful tuning and may fail in dynamic scenarios \mycite{macenski2023from}.
Moreover, while controller stability can emerge from configuring sampling and cost design, neither stability nor safety can be guaranteed.

\subsection{Model Predictive Path Integral (MPPI)}
\label{subsec:mppi}

\ac{mppi} \mycite{williams2016aggressive}, a model-predictive controller and a specialization of reactive controllers, optimizes a cost function online by sampling a search space.
Extensions to \ac{mppi} introduce constraints during sampling to provide safety guaratees~\mycite{borquez2025dualguard}.
However, adaptivity remains limited by static, risk-neutral sampling, motivating uncertainty-based extensions~\mycite{mohamed2025toward}.

% \ac{mppi} \mycite{williams2016aggressive}, a model-predictive controller and a specialization of reactive controllers, optimizes a cost function online by sampling a search space.
% This allows it to not only consider circular trajectories (as done by \ac{dwb}) but also enables timely reactions to changing environments.
% While not implemented in its original form, extensions to \ac{mppi} have been proposed to provably guarantee safety using constraintes during trajectory sampling~\mycite{borquez2025dualguard}.
% On the other hand, adaptivity to dynamic environments is limited by the static, risk-neutral trajectory sampling and motivated to be extended using uncertainty for efficient trajectory sampling in~\mycite{mohamed2025toward}.

\subsection{\ac{rl} in Path Tracking}
\label{subsec:rl}

\ac{rl} enables agents, that is, training-based controllers to learn from interactions with their environment, adjusting their actions based on rewards and penalties they receive.
By maximizing cumulative rewards over time, \ac{rl} agents can learn optimal control policies without requiring explicit programming or supervision.
Learning can be done online, where the agent learns while interacting with the environment, or offline, where the agent learns from a dataset of past interactions.
Thus, even in complex and dynamic environments, \ac{rl}-based controllers can learn optimal control policies that adapt to changing conditions~\mycite{singh2022reinforcement}.

Recent studies have explored various ways to apply \ac{rl} to path tracking.
These include using a path-integral based \ac{rl} approach to learn optimal parameters for a path following controller offline and tune them online with real-time system states, leading to improved performance and adaptability~\mycite{zhu2020path}.
A hierarchical approach has also been proposed by~\mycite{chai2024design}, where a high-level controller plans a path based on a global map and a low-level controller uses \ac{rl} to track the path while avoiding obstacles in local, unknown environments.
Q-learning has been combined with PID control to improve tracking accuracy and reduce computational complexity in~\mycite{wang2020trajectory}.
Another approach is proposed by~\mycite{luy2014reinforcement}.
They developed an integrated kinematic and dynamic tracking control algorithm based on \ac{rl} that does not require knowledge of the robot's dynamic model, leading to improved robustness and adaptability~\mycite{luy2014reinforcement}.
Furthermore, their approach is the only one of the listed that provides guarantees on the stability of the closed-loop system.
As such, while \ac{rl} approaches show potential in handling complex path tracking situations and despite emerging approaches in safe \ac{rl} \mycite{garcia2015comprehensive}, \ac{rl} in general still falls short in providing guarantees on safety or stability.

\subsection{Discussion}
\label{subsec:discussion}

Comparing traditional controllers readily implemented, e.g., in \ac{ros2}, with recent \ac{rl} approaches, highlights the divergence between adaptability and safety.
Fundamental methods such as \ac{rpp} and \ac{mppi} provide stability guarantees with acceptable but, at times, limited performance.
This raises the question of adaptivity to increase performance in complex, dynamic environments, which commonly comes at the cost of safety and stability guarantees -- even considering extensions of these foundational methods.
The situation is aggrevated by \ac{rl}-based controllers, which promise adaptability and performance in complex scenarios but lack guarantees on safety or stability due to their black-box nature and learning-based design.
Thus, the need for adaptivity in path tracking produces controllers that cannot be deployed in safety-critical applications.
However, is there a way to combine the strengths of both traditional and \ac{rl}-based controllers, ensuring safety while leveraging adaptability and performance?

%%%%
%% Related Work
%%%%
\section{The Simplex Architecture}
\label{sec:related_work}
Hybrid architectures combine different controllers to exploit their individual characteristics.
In this section, we examing how the Simplex architecture~\mycite{sha2001using} can be used to combine the adaptability of high-performance controllers (e.g., \ac{rl}-based) with the safety guarantees of traditional controllers.
We first introduce its structure and components (\cmpr{subsec:simplex_architecture}), then formulate our contribution: design considerations for stability and safety guarantees in the Simplex architecture (\cmpr{subsec:stability_analysis}).

% Hybrid architectures aim to combine different controllers to exploit their individual characteristics.
% To address the question of how to combine adaptability of high-performance controllers (e.g., \ac{rl}-based), with the safety guarantees of traditional controllers, we discuss the Simplex architecture~\mycite{sha2001using} in this section.
% After introducing its structure and components (\cmpr{subsec:simplex_architecture}), we formulate our first contribution: design considerations for providing stability and safety guarantees in the context of the Simplex architecture (\cmpr{subsec:stability_analysis}).

\subsection{Components of the Simplex Architecture}
\label{subsec:simplex_architecture}

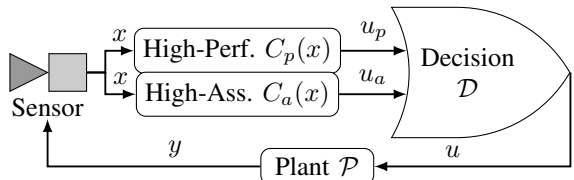
\begin{figure}
	\centering
	\begin{tikzpicture}[circuit logic CDH]
		\node[arcnode, minimum width=1.5cm] (p) at (1.25,-0.75) {Plant $\plant$};
		%\node[arcnode, rotate=90, minimum width=1.5cm] (d) at (2,0.5) {Decision $\decision$};
		\node[or gate US, draw,logic gate inputs=pp, text width=1.5cm, align=center, fill=white] (d) at (3.25,0.5) {Decision $\decision$};

		\node[arcnode, minimum width=1.75cm] (c) at ($(d.input 1) + (-2.25,0)$) {High-Perf. $C_p(x)$};
		\node[arcnode, minimum width=1.75cm] (s) at ($(d.input 2) + (-2.25,0)$) {High-Ass. $C_a(x)$};

		\Sensor[90][2.5][Sensor]{$(c.west)!0.5!(s.west) + (-0.25*\shift,0)$}
		\node[anchor=north] (SensorLabel) at ($(Sensor.west) + (0.0, 0.05*\shift)$){Sensor};

		\draw[flowedge] (c.east) -- node[above] {$u_{p}$} (d.input 1);
		\draw[flowedge] (s.east) -- node[above] {$u_{a}$} (d.input 2);
		\draw[flowedge] (d.output) |- node[above, pos=0.8] {$u$} (p.east);
		\draw[flowedge] (p.west) -| node[above, pos=0.2] {$y$} ($(Sensor.west) + (0,-0.5)$) -- ($(Sensor.west)+(0,-0.35)$);
		\draw[flowedge] (Sensor.south) -- ([xshift=0.25cm]Sensor.south) |- node[above,pos=0.75] {$x$} (c.west);
		\draw[flowedge] (Sensor.south) -- ([xshift=0.25cm]Sensor.south) |- node[above,pos=0.75] {$x$} (s.west);
	\end{tikzpicture}
	\caption{The Simplex architecture~\mycite{sha2001using} combines a high-performance, adaptive controller with a high-assurance, fallback controller that is switched to if the decision module detects unsafe actions.}
	\label{fig:simplex_architecture}
\end{figure}
The Simplex Architecture proposed by \mycite{sha2001using} combines controllers to achieve safety while leveraging the performance advantages of less verifiable controllers.
For that, the architecture consists of three main components: a high-performance controller $C_p(x)$, a high-assurance controller $C_a(x)$, and a decision module $\decision$ (with $x \in \mathcal{X} \subseteq \mathbb{R}^n$ being the state of the system) as illustrated in \cref{fig:simplex_architecture}.
Together, they realize a separation of concerns.
The high-performance controller is designed for optimal performance, may be complex and adaptive, and is not required to provide guarantees.
The high-assurance controller is designed to stabilize the system and guarantee safety, potentially at the cost of performance.
The decision module acts as a switching mechanism between the two controllers, monitoring the system's state and the actions proposed by the high-performance controller.
It can perform two actions:
\begin{enumerate*}[label=(\arabic*)]
	\item switch to the high-assurance controller if the high-performance controller's actions may lead to unsafe behavior, and
	\item switch back to the high-performance controller if the system is safe and stable.
\end{enumerate*}
Thus, while the high-assurance controller and decision module are concerned with safety and stability, the high-performance controller focuses on performance and adaptability.

\subsection{Design Considerations for Stability and Safety}
\label{subsec:stability_analysis}

The separation of concerns in the Simplex architecture provides a framework to achieve safety and stability but requires careful design of the high-assurance controller and the decision module.
For that, the difference between safety and stability needs to be clarified.
Stability refers to the system's ability to return to a stable state after a disturbance, while safety refers to the system's ability to minimize the risk of hazardous events by preventing entering associated states (from which it still might be able to stabilize).
Thus, while stability is a necessary condition for safety, it is not sufficient.
\gj{can we find a better example than this? We consider an minimal error to be sufficient for safety in our case, so this is not consistent. Can we find an better, unrelated example?}
For instance, a robot may stably follow a path but still collide with an obstacle due to unsafe oscillations.

In this line, we discuss design considerations for the high-assurance controller to provide stability guarantees and the decision module to implement a strategy for switching between the high-performance and high-assurance controller to maintain system safety.
We first discuss stability as a basis of safety in \cref{subsubsec:stability} to derive stability guarantees for the high-assurance controller.
Then, we integrate safety constraints to derive a strategy for the decision module to switch to the high-assurance controller when needed in \cref{subsubsec:integrating_safety}.
Finally, we discuss the concept of dwell time~\mycite{cao2010dwell,xiang2022necessary} in \cref{subsubsec:dwell_time} to derive a strategy for safely switching back to the high-performance controller.

\subsubsection{Stability Considerations}
\label{subsubsec:stability}

A necessary condition for safety is stability.
As the high-performance controller $C_p(x)$ does not provide stability guarantees and may even destabilize the system, the high-assurance controller $C_a(x)$ needs to ensure stability.
Moreover, the assumptions and limitations tied to these guarantees (e.g., as arising through linearization) need to be explicitly known by the decision module since it can only rely on the high-assurance controller to stabilize the system when the assumptions hold.
In other words, the decision module needs to switch to the high-assurance controller before control actions of the high-performance controller $C_p(x)$ drive the system to states where the stability guarantees of $C_a(x)$ are no longer valid.
Thus, stability for the Simplex architecture serves two purposes: ensuring the high-assurance controller can stabilize the system and providing a criterion for the decision module to switch to the high-assurance controller.

A concept applicable here is the \ac{roa}, defining the region $\mathcal{X}_{A|C} \subseteq \mathcal{X}$ within the state space where a controller $C(x)$ guarantees system stabilization.
The \ac{roa} can be determined through stability analysis methods like Lyapunov functions~\mycite{liberzon1999basic}, incremental stability~\mycite{bernardo2016switching}, or reachability analysis~\mycite{althoff2021set}.
These approaches are suitable for non-linear systems as is the case for path tracking.

Lyapunov functions are scalar functions that decrease along system trajectories that converge towards an equilibrium point, providing a measure of the system's energy~\mycite{liberzon1999basic}.
Thus, a system is stable if it minimizes the Lyapunov function.
Its region of attraction is then defined as the level set of the Lyapunov function.
However, finding a Lyapunov function for an arbitrary system can be challenging and leads to conservative approximations of the actual \ac{roa}~\mycite{liberzon1999basic}.

Incremental stability focuses on the convergence of any two system trajectories, providing a measure of stability even without a specific equilibrium point~\mycite{bernardo2016switching}.
Assuming that a set of states $\mathcal{X}_I$ can be found for which the system's trajectories start and end in $\mathcal{X}_I$ while converging over time, the set is said to be forward invariant and can be interpreted as a \ac{roa} ($\mathcal{X}_I = \mathcal{X}_A|C$).
However, defining this forward invariant set explicitly is challenging for complex systems but needed for deriving a switching strategy for the decision module.

Reachability analysis aims to determine (all) possible system states that can be reached from a given initial state, providing a comprehensive view of the system's behavior and directly resulting in a \ac{roa}~\mycite{althoff2021set}.
Approaches for calculating reachable sets can be hybridization-based, Taylor-model-based, constraint-solving-based, or simulation-based~\mycite{dongxu2020reachability}.
\mycite{althoff2021set} propose set propagation techniques, which allow for the analysis of hybrid systems.

With these methods, stability of high-assurance controllers can be analyzed and their \ac{roa} determined.
This region provides stability guarantees and serves as a basis for designing the decision module's condition to switch to the high-assurance controller.

\subsubsection{Integrating Safety Constraints}
\label{subsubsec:integrating_safety}

While stability is a necessary condition for safety, it is not sufficient.
Safety requires minimizing risks beyond stability, such as avoiding collisions or staying within specified boundaries.

The approach of \ac{cbf} offers a powerful mechanism to encode safety requirements during controller synthesis or verification~\mycite{wang2025safe}.
They are positive-definite functions defined on the system's state space, typically decreasing towards the desired equilibrium point.
Safety requirements are mapped to a constant threshold value that the \ac{cbf} must not exceed.
This approach enables handling safety as inequality constraints during controller synthesis, effectively combining safety and stability considerations within the region of attraction.

Alternatively, barrier states, that is, states to be avoided by the system, can be defined~\mycite{almubarak2022safety}.
This method utilizes \ac{cbf}s as new state variables embedded within the system's state space.
While this allows applying standard stability analysis methods without explicitly considering safety constraints, it can increase system complexity and potentially render the system uncontrollable.

For defining the switching condition of the decision module, however, controller synthesis is not required.
Thus, an intuitive approach is to restrict the region of attraction of the high-assurance controller based on the safety requirements, that is, $\mathcal{X}_{S|C} = \mathcal{X}_{A|C} \setminus \left\{ x \mid s(x) > 0 \forall x \in \mathcal{X}, \forall s \in \mathcal{S} \right\}$, where $\mathcal{S}$ is the set of safety constraints.
By additionally requiring convexity of the set and ensuring invariance, the region $\mathcal{X}_{S|C}$ represents the safe operating space under both stability and safety constraints.
The decision module can then monitor the system state relative to the boundary of $\mathcal{X}_{S|C}$ and switch to the high-assurance controller when the system is about to enter a state $x'_{C_p} \notin \mathcal{X}_{S|C}$, where $x'_{C_p}$ is the one-step predicted state based on the high-performance controller's actions and the overall control period.

\subsubsection{Dwell Time and Stability of Switching Systems}
\label{subsubsec:dwell_time}

With the switch from high-performance to high-assurance controller being defined, the question arises when to switch back to the high-performance controller.
For that, the decision module needs to ensure that the system is stabilized under the high-assurance controller before switching back.

A concept that arose from the stability analysis of switched systems is the (average) dwell time, denoted as $T_d$~\mycite{cao2010dwell,xiang2022necessary}.
It represents the minimum duration a system needs to stabilize under a given controller before switching to another to avoid oscillations between controllers that could lead to instability (despite individual controllers being stable).
The concept of average dwell time is a generalization of dwell time that allows for some switches to occur more quickly than the minimum dwell time, as long as the average time between switches is sufficiently long.

Since the Simplex architecture constitutes a switching system, the average dwell time can be used to determine when it is safe to switch back to the high-performance controller.
It constitutes a simplistic yet effective approach to ensure stability.
With that, the decision module's function can be defined as\footnote{Note that we simplify notation by omitting the time, i.e., $x = x(t)$, $u = u(t)$, etc.}:
\begin{equation}
	\decision(x) = \begin{cases}
		C_p(x) & \text{if } x'_{C_p} \in \mathcal{X}_{S|C} \text{ and } t \geq t_{HA} + T_d \\
		C_a(x) & \text{otherwise}
	\end{cases}
	\label{eq:decision_module}
\end{equation}
where $t_{HA}$ is the time when the system switched to the high-assurance controller.
This ensures that the system is stabilized under the high-assurance controller before switching back to the high-performance controller.

%%%%
%% Concept
%%%%
\section{Simplex-based Path Tracking Controller}
\label{sec:concept}
\begin{figure*}[!ht]
	\centering
	\begin{subfigure}{0.6\textwidth}
		\centering
		\includegraphics[width=0.9\linewidth]{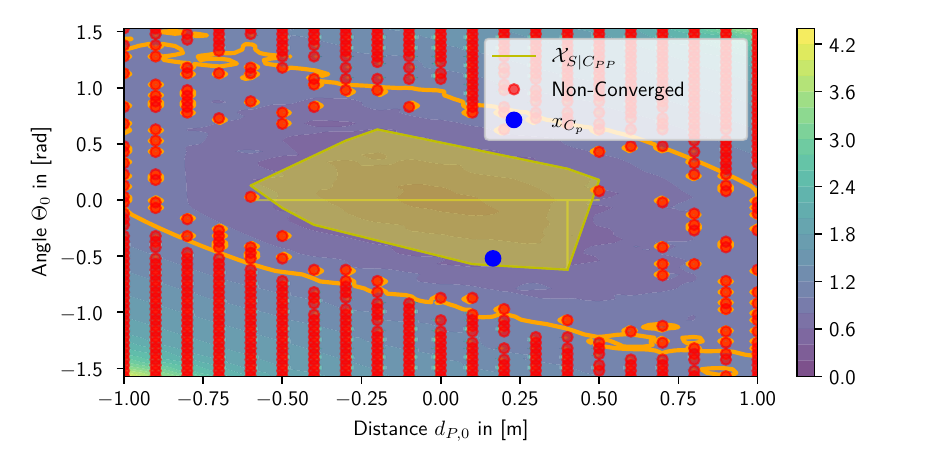}
		\caption{Contour plot of maximal deviation from the path for different starting positions under $C_{PP}$.\\}
		\label{fig:max_distance_max}
	\end{subfigure}
	\begin{subfigure}{0.3\textwidth}
		\centering
		\resizebox{0.9\linewidth}{!}{%
			\input{figures/generated/path_trajectory.tex}
		}
		\caption{Single simulation of reachability analysis with a starting state $d_{\mathcal{P},0} = \SI{0.5}{\meter}$, $\Theta_0 = \SI{0.0792}{\radian}$, $r_p = 0.8$.}
		\label{fig:ha_sim}
	\end{subfigure}
	\caption{Results of reachability analysis.}
	\label{fig:ra}
	\vspace{-0.25cm}
\end{figure*}

With the last section's discussion, we now apply the Simplex architecture to design a path tracking controller for a differential drive mobile robot.
We start with the system definition (\cref{subsec:system_definition}), then define the controllers (\cmpr{subsec:definition_of_controllers}), and finally describe a procedure to derive a safe switching strategy based on the region of attraction of the high-assurance controller and the safety requirements (\cmpr{subsec:decision_module}).

\subsection{System Definition}
\label{subsec:system_definition}

For the path tracking problem, we consider a differential drive mobile robot with nonlinear kinematics:
\begin{equation}
	\begin{bmatrix}
		\dot{x} \\
		\dot{y} \\
		\dot{\theta}
	\end{bmatrix} =
	\begin{bmatrix}
		\cos(\theta) & 0 \\
		\sin(\theta) & 0 \\
		0            & 1
	\end{bmatrix}
	\begin{bmatrix}
		v \\
		\omega
	\end{bmatrix}
	\label{eq:kinematics} % Place the label command here
\end{equation}
where $(x,y)$ represents the robot's position in the 2D plane, $\theta$ is its orientation, and $v$ and $\omega$ are the linear and angular velocities given as control actions by the controller $C(x)$, which we want to design using the Simplex architecture.
The path tracking problem requires guiding the robot along a given path $\mathcal{P} = \{p_0, \ldots, p_m\}$ with $p_j \in \mathbb{R}^2$, defined as a sequence of waypoints $p_j$ in the 2D plane.
For that, $C(x)$ is limited to the maximal velocity $v \in [0, v_{\text{max}}]$ with $v_{\text{max}} = \SI{1}{\meter\per\sec}$ and angular velocity $\omega \in [-\omega_{\text{max}}, \omega_{\text{max}}]$ with $\omega_{\text{max}} = \SI{0.5}{\radian\per\sec}$ originating from the robot's physical constraints.
Additionally, $C(x)$ must adhere to safety requirements $\mathcal{S}$ during path tracking.
Motivated by the real-world example of a delivery scenario (\cmpr{fig:robot_sidewalk}) where the robot has to drive on a sidewalk with a width of approx. $\SI{2}{\meter}$, we set the safety margin to $\SI{1}{\meter}$ which results in $\mathcal{S} = \left\{s_{\mathcal{P}}(x) = d_{\mathcal{P}}(x) \leq  \SI{1}{\meter}\right\}$, where $d_{\mathcal{P}}(x)$ denotes the robot's distance to the path.

\subsection{Definition of Controllers}
\label{subsec:definition_of_controllers}

Given the system definition and the control task, we can define the controllers to be used in the Simplex architecture.
We describe the \ac{rl}-based, high-performance controller next and detail the high-assurance controller afterward.

\subsubsection{High-Performance Controller}
\label{subsubsec:high_performance_controller}

Since no guarantees on safety or stability are required for the high-performance controller, any controller can be used.
Here, we implement an \ac{rl}-based high-performance controller using the \ac{td3bc} algorithm~\mycite{fujimoto2021minimalist}.
It is an offline \ac{rl} algorithm, enabling us to leverage existing data generated in a simulated environment and eliminating the need for extensive real-world training.
Specifically, the behavior cloning component allows us to incorporate expert knowledge from existing controllers, further enhancing the learning process.
We exploited this by deploying the \ac{mppi}, \ac{rpp}, \ac{dwb} controllers for various waypoint following tasks using Gazebo and \ac{ros2}~\mycite{macenski2022robot} to generate a dataset covering diverse scenarios.
By applying \ac{td3bc}, we trained a high-performance controller $C_{RL}(x)$.
Additionally, we trained a suboptimal controller $\hat{C}_{RL}(x)$ by prematurely terminating the training process to induce unsafe control actions, which we use for evaluating the Simplex architecture's ability to prevent safety violations.

\subsubsection{High-Assurance Controller}
\label{subsubsec:high_assurance_controller}

For the high-assurance controller, which is required to provide stability guarantees, we exploit the results of \mycite{ollero95stability} and implement $C_{PP}(x)$ using the pure pursuit algorithm.
We configure the controller with a lookahead distance of $L=\SI{1.0}{\meter}$ and a constrained linear velocity of $v_{max, PP} =\SI{0.5}{\meter\per\sec}$ to ensure adherence to the safety requirements in $\mathcal{S}$.
Following our discussion in \cref{subsec:stability_analysis}, we approximate the \ac{roa} of $C_{PP}(x)$ using simulation-based reachability analysis.
We solve the kinematic model using Runge-Kutta 4/5 with a step size of $\SI{0.05}{\sec}$ for $T=\SI{15}{\sec}$, for 11907 different initial states.
These states vary in initial distance to the path ($d_{\mathcal{P},0} \in [-\SI{1.0}{\meter},\SI{1.0}{\meter}]$ with step size $\SI{0.1}{\meter}$), initial orientation relative to the path ($\Theta_0 \in [-\SI{1.5708}{\radian},\SI{1.5708}{\radian}]$ with step size $\SI{0.05}{\radian}$), and the robot's relative position between the first two waypoints ($r_p \in [\SI{0}{\meter}, \SI{0.4}{\meter}]$ with step size $\SI{0.05}{\meter}$).
We generate 100 different, random paths with 50 waypoints each, resulting in 1190700 simulations.
The system is considered stabilized if the robot converges to the path with a distance of $d_{\mathcal{P}} \leq \SI{0.1}{\meter}$ for at least $\tau=\SI{1.5}{\sec}$.
For each simulation, we recorded the maximum distance to the path $\max(d_{\mathcal{P}})$ and the convergence time $T_c$.

\cref{fig:max_distance_max} shows the maximum deviation $\max(d_{\mathcal{P}})$ over all simulations as a contour plot depending on $d_{\mathcal{P},0}$ and $\Theta_0$ as obtained from the reachability analysis.
Red dots denote initial states resulting in non-converging simulations, i.e., where $C_{PP}(x)$ could not stabilize the robot.
The orange line indicates the level set where $d_{\mathcal{P}}=\SI{1.0}{\meter}$ constituting our safety requirement.
As non-converging states can be found within this level set, however, the \ac{roa} $\mathcal{X}_{A|C_{PP}}$ is limited by them.
The simulation associated with the non-converging state closest to $\mathcal{X}_{A|C_{PP}}$ is shown in \cref{fig:ha_sim}, where the $C_{PP}$ limitation of having a fixed lookahead distance of $L=\SI{1.0}{\meter}$ is visible.
The robot's initial distance to the path paired with its randomness in curvature and the fixed lookahead distance can lead to oscillating behavior violating our convergence criteria.
Although oscillations are decreasing, the controller's \ac{roa} is restricted by them.

Nevertheless, the results reinforce the stability guarantees reported by \mycite{ollero95stability}, which assume closeness to the path.
This analysis provides the basis for the decision module to derive the switching strategy, which we discuss in the next section.

\subsection{Decision Module}
\label{subsec:decision_module}

Following the discussion in \cref{subsec:stability_analysis} and specifically \cref{eq:decision_module}, we need to define two switches for the decision module.
The first switch is from the high-performance controller to the high-assurance controller when the system's safety is in danger.
The second switch is back to the high-performance controller when the system is safe and stabilized again.

%% from HP to HA
For the first switch, we consider the \ac{roa} $\mathcal{X}_{A|C_{PP}}$ of the previous subsection, which has to be intersected with the safety requirements $\mathcal{S}$ stating that the robot cannot diverge more than $\SI{1.0}{\meter}$ from the path.
For that, we consider the maximum deviation recorded during the simulation for each state in $\mathcal{X}_{A|C_{PP}}$, which is $\max_{\mathcal{X}_{A|C_{PP}}}(d_{\mathcal{P}}) = \SI{0.681}{\meter}$.
This is caused by the already discussed non-converging state at $(d_{\mathcal{P},0} = \SI{0.5}{\meter},\Theta_0 = \SI{0.0792}{\radian})$ that limits the \ac{roa}.
However, since this means that the maximal deviation that might be encountered when switched to the high-assurance controller for stabilizing the system is well within the safety requirement, $\mathcal{X}_{A|C_{PP}} = \mathcal{X}_{S|C_{PP}}$ in our case.
Thus, both sets are indicated by the yellow area/line in \cref{fig:max_distance_max}. % indicates both sets, $\mathcal{X}_{A|C_{PP}}$ and $\mathcal{X}_{S|C_{PP}}$.

%% explaining the reduced set for simplifying the decision module
At this point, we simplify the formulation of the decision module.
According to \cref{eq:decision_module}, the decision module switches depending on the predicted state $x'_{C_{p}}$.
To avoid implementing an online simulation, we instead further restrict $\mathcal{X}_{S|C_{PP}}$.
Given the maximal deviation from the path encountered during simulations of states in the \ac{roa} is $\max_{\mathcal{X}_{A|C_{PP}}}(d_{\mathcal{P}}) = \SI{0.681}{\meter}$ and given the robot's maximal velocity $v_{\text{max}} = \SI{1}{\meter\per\sec}$, the robot can move at most $\SI{0.05}{\meter}$ within one control period.
We use this to restrict the maximal deviation from the path $\max_{\mathcal{X}_{A|C_{PP}}}(d_{\mathcal{P}}) = \SI{0.681}{\meter} - \SI{0.05}{\meter} = \SI{0.631}{\meter}$ and form $\mathcal{X}'_{S|C_{PP}}$ as a \ac{roa}.
Now, we can use $x_{C_p} \notin \mathcal{X}'_{S|C_{PP}}$ as a condition for the decision module to switch to the high-assurance controller.
Since we know that the robot can move at most $\SI{0.05}{\meter}$ within one control period, the robot will still be in $\mathcal{X}'_{S|C_{PP}}$ when triggering the switch while not being in $\mathcal{X}'_{S|C_{PP}}$ anymore.
An example (blue point) is shown in \cref{fig:max_distance_max} where the decision module activated the high-assurance controller while the system remained within the original $\mathcal{X}_{S|C_{PP}}$.
This is further discussed in the evaluation section, together with \cref{fig:simulation_cosine}.

%% from HA to HP
For the second switch, we consider the dwell time of the high-assurance controller, \cmpr{subsubsec:dwell_time}.
For each state in $\mathcal{X}'_{S|C_{PP}}$, we recorded the time needed to converge to the path $T_c$.
Assuming the worst-case again, the maximum time needed to converge to the path is $\max_{T_c \in \mathcal{X}'_{S|C_{PP}}}(T_c) = T_d = \SI{12.45}{\sec}$.
Using the parameters $T_d$ and $\mathcal{X}'_{S|C_{PP}}$ together with $C_{PP}(x)$ and $C_{RL}(x)$, the decision module as defined by \cref{eq:decision_module} is parameterized and thereby constitutes our Simplex architecture for path tracking, which we denote as $C_S(x)$ for simplicity.
Note, for evaluation purposes, we constructed the Simplex controller using the suboptimal controller $\hat{C}_{RL}(x)$ as well and denote the resulting Simplex controller as $\hat{C}_S(x)$.

%%%%
%% Evaluation
%%%%
\vspace{-0.1625cm}
\section{Evaluation}
\label{sec:evaluation}

\gj{Idea for evaluation: fault injection to steer the introduced sensor noise. Drive the robot along a given GPS path several times to find the baseline - let the robot record its own perception of distance to the path. Then, inject noise to the GPS signal and evaluate the performance of the controller (average speed, mean crosstrack error, maximal distance to path). How often does the Simplex controller need to switch to the high-assurance controller? How does the performance of the Simplex controller compare to the performance of the high-performance controller in isolation?}

We evaluate the designed and parameterized Simplex controllers $C_S$ and $\hat{C}_{S}$ in this section by means of extended simulation and comparison to other controllers, including \ac{mppi}, \ac{rpp}, and \ac{dwb}. %as well as the controllers $C_{PP}$, $\hat{C}_{RL}$, and $C_{RL}$ used within $\hat{C}_S$ and $C_S$.
We present our simulation setup in the next subsection before discussing the results and comparing the Simplex controller to other controllers.

\subsection{ROS2 Simulation Setup}
\label{subsec:ros2_imp}
%%% Squared Track
\begin{figure}
	\renewcommand\thesubfigure{\alph{subfigure}}
	\centering
	\begin{subfigure}{0.45\textwidth}
		\centering
		\includegraphics[width=0.85\linewidth]{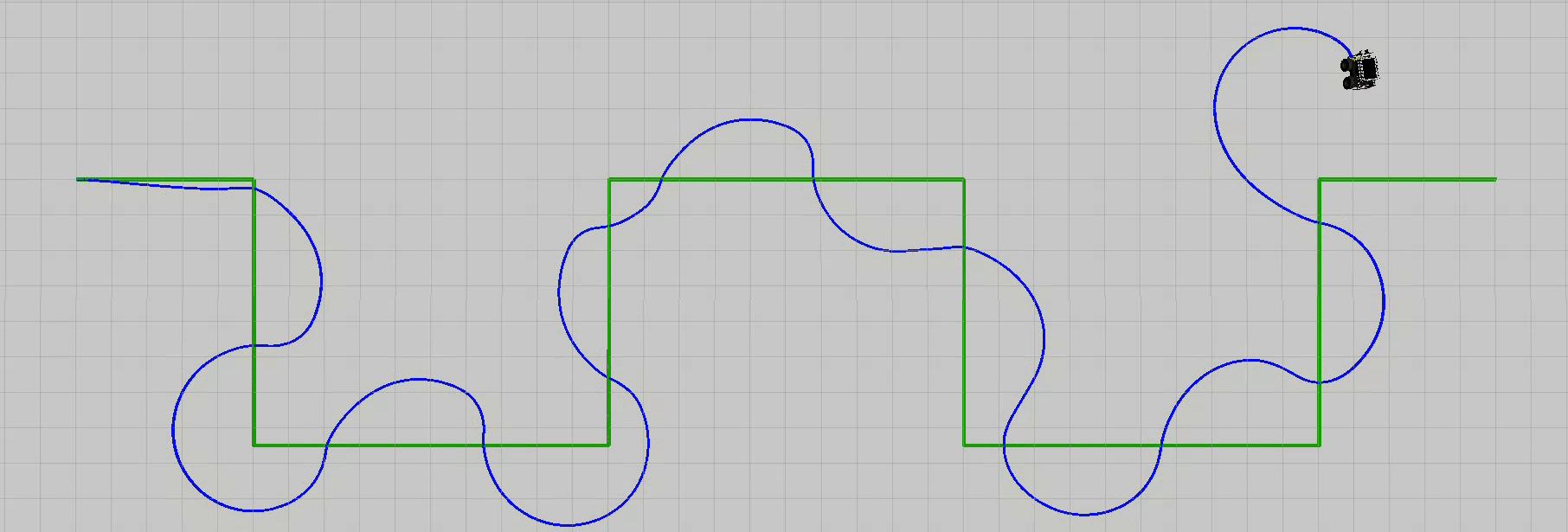}
		\caption{Unsafe $\hat{C}_{RL}$ controller failing safety requirement.}
		\label{fig:pytorch_failing}
	\end{subfigure}
	\begin{subfigure}{0.45\textwidth}
		\centering
		\includegraphics[width=0.85\linewidth]{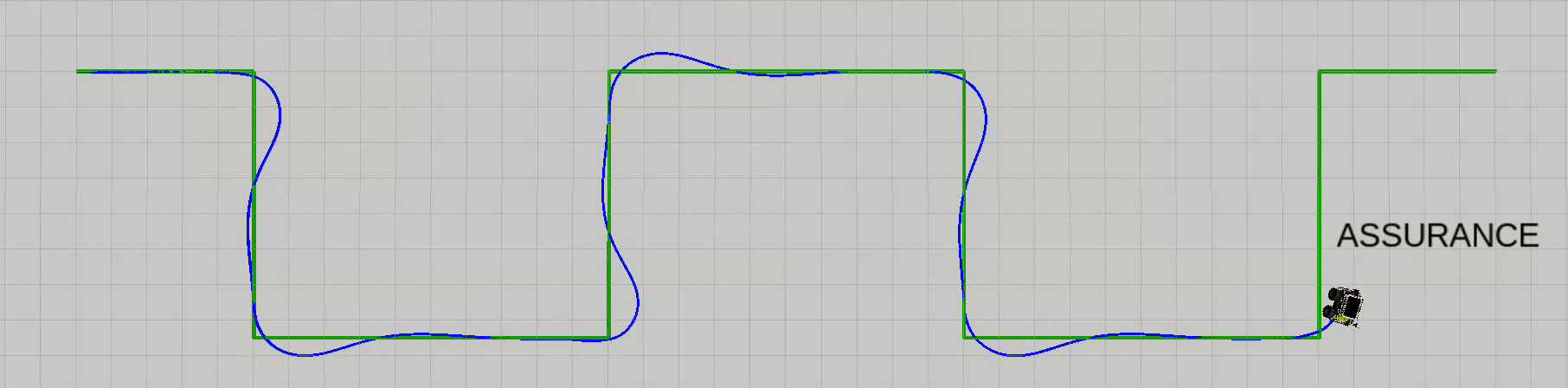}
		\caption{$\hat{C}_S$ maintains safety through high-assurance controller.}
		\label{fig:simplex_assurance}
	\end{subfigure}
	\caption{Comparing $\hat{C}_{RL}$ and $\hat{C}_S$ controllers on $\mathcal{P}_{square}$ track.}
	\label{fig:simulation}
	\vspace{-0.25cm}
\end{figure}
%%% Cosine Track
\begin{figure}
	\centering
	\begin{subfigure}{0.45\textwidth}
		\centering
		\includegraphics[width=0.85\linewidth]{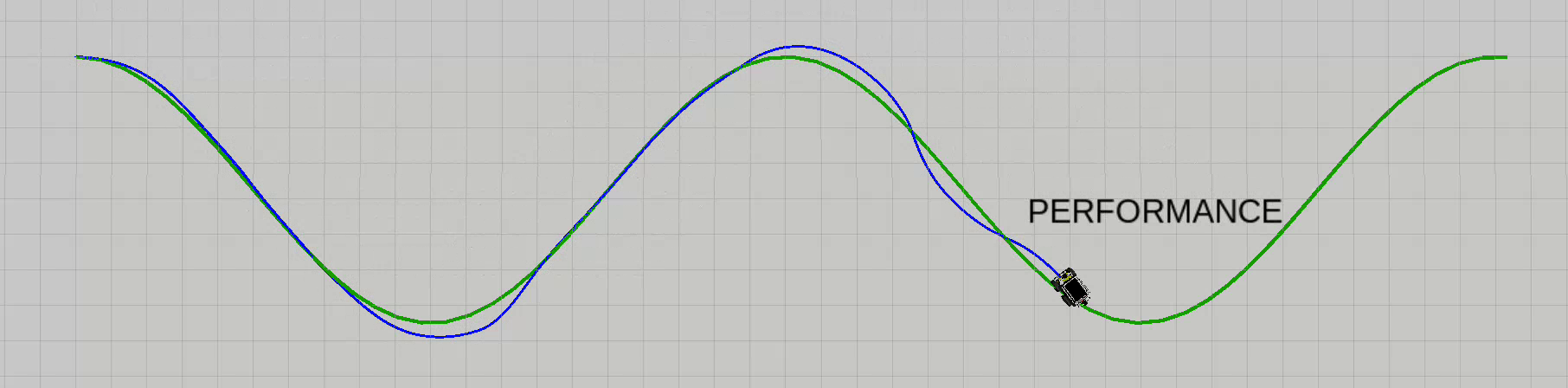}
		\caption{$\hat{C}_S{x}$ in high-performance mode at $t=\SI{69}{\sec}$.}
		\label{fig:cosine_hp_mode}
	\end{subfigure}
	\begin{subfigure}{0.45\textwidth}
		\centering
		\includegraphics[width=0.85\linewidth]{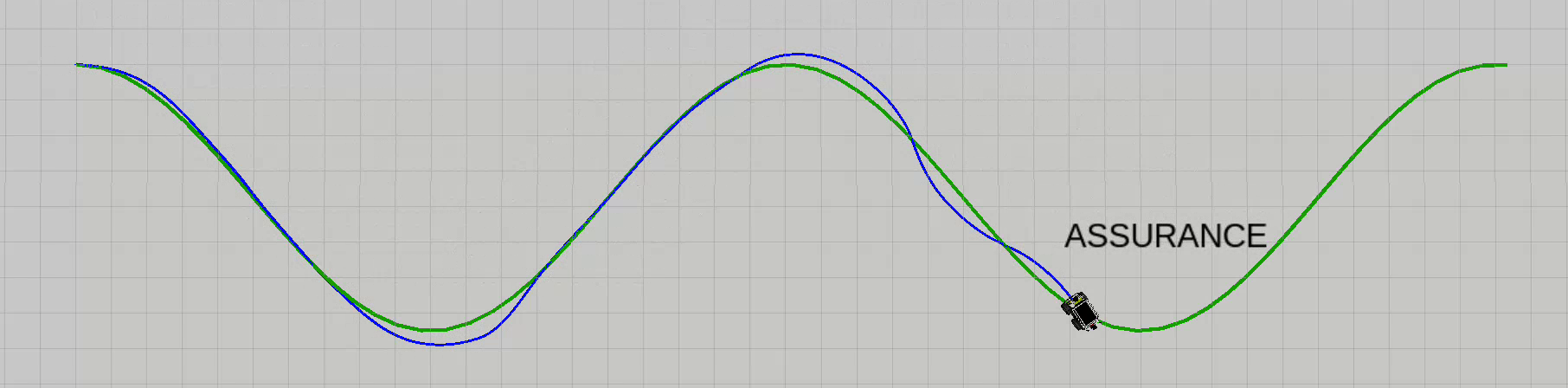}
		\caption{$\hat{C}_S{x}$ in high-assurance mode at $t=\SI{70}{\sec}$.}
		\label{fig:cosine_ha_mode}
	\end{subfigure}
	\caption{Switch from high-performance (\cmpr{fig:cosine_hp_mode}) to high-assurance mode (\cmpr{fig:cosine_ha_mode}) in the Simplex controller due to relative angle ($\Theta_0 = \SI{-0.52}{\radian}$) between robot and path. The state $x_{C_p}$ is visualized as a blue point in \cref{fig:max_distance_max}.} % ($\Theta_0 = \SI{-0.52}{\rad}$, $d_{\mathcal{P}} = 0.165$)
	\label{fig:simulation_cosine}
	\vspace{-0.25cm}
\end{figure}
%% Simulation Environment
To evaluate the controllers, we use the Gazebo\footnote{\url{https://gazebosim.org/home}, Accessed: 2025-09-08} simulation environment where we modeled the Clearpath Husky robot\footnote{\url{https://clearpathrobotics.com/husky-a300-unmanned-ground-vehicle-robot/}, Accessed: 2025-09-08} as shown in \cref{fig:robot_sidewalk}.
For implementing the controllers, we use the \ac{ros2} navigation stack and the existing implementations of the \ac{mppi}, \ac{rpp}, and \ac{dwb} controllers and leverage the plugin-based structure to implement our own controllers ($C_{PP}$, $\hat{C}_{RL}$, $C_{RL}$, $\hat{C}_{S}$, $C_S$).

%% Tracks
For defining the path to be followed by the robot, we consider two different tracks: a squared track $\mathcal{P}_{square}$ and a curved track $\mathcal{P}_{cos}$.
The squared track is motivated by the delivery scenario in urban environments with sharp turns at intersections, \cmpr{fig:simulation}.
The track is $\SI{70}{\meter}$ long in total, with each straight segment being $\SI{10}{\meter}$ (sideways) and $\SI{7.5}{\meter}$ (downwards) long.
Additionally, we considered a curved track given as a cosine function (\cmpr{fig:simulation_cosine}) for assessing the controllers' performance capabilities in a simplified environment.
%% Execution
Each controller was simulated on each track for $N=30$ times, resulting in a total of $N=480$ simulations.

\subsection{Results}
\label{subsec:results}

This section discusses the simulation results.
First we exemplarily analyze the Simplex controller's safety assurance through \cmpr{fig:simulation} and \cmpr{fig:simulation_cosine} and then compare it to other state-of-the-art path tracking algorithms in \cref{tab:performance}.

\subsubsection{Safety Assurance}
\label{subsubsec:safety}
%The Simplex architecture enables the use of complex, high-performance controllers (e.g. $\hat{C}_{RL}(x)$) in safety-critical applications by ensuring safety through a high-assurance controller ($C_{PP}(x)$).
To analyze the safety assurance achieved through the Simplex controller, we simulated the $\hat{C}_{RL}(x)$ controller in isolation and embedded in $\hat{C}_S(x)$ on the squared track $\mathcal{P}_{square}$.
\Cref{fig:simulation} shows the exemplary simulations visualized using RViz2\footnote{\url{https://github.com/ros2/rviz}, Accessed: 2025-09-08}.
The green line represents the planned path $\mathcal{P}_{square}$ and the blue line the robot's actual trajectory.

In \cref{fig:pytorch_failing}, $\hat{C}_{RL}(x)$ is failing to maintain a safe distance to the path, a maximal deviation of $\max(d_{\mathcal{P}}) = \SI{2.753}{\meter}$ is reached over all our simulations, \cmpr{tab:performance}.
Especially the sharp turns are challenging for the controller.
In \cref{fig:simplex_assurance}, the decision module detects that $x \notin \mathcal{X}'_{S|C_{PP}}$ and switches to the high-assurance controller $C_{PP}(x)$.
The mode is indicated in the figure through the \textit{ASSURANCE} label.

A similar situation is observed for \cref{fig:simulation_cosine} where we display the switching from high-performance to high-assurance mode in the Simplex controller due to the relative angle ($\Theta_0$) between robot and path.
Being in high-performance mode on the (almost) straight segment of the track (\cmpr{fig:cosine_hp_mode}), $\hat{C}_{RL}$ oscillates and causes the robot to cross the path such that the relative angle between robot and path exceeds the threshold, that is, $x_{C_p} \notin \mathcal{X}'_{S|C_{PP}}$ (\cmpr{fig:max_distance_max}).
Thus, in the next step, the Simplex controller switches to high-assurance mode (\cmpr{fig:cosine_ha_mode}) to ensure safety.

\subsubsection{Comparison to Other Controllers}
\label{subsubsec:comparison}
To further support the safety assurance provided by the Simplex controller, we compare its performance to other controllers.
For that, \cref{tab:performance} shows the average tracking error $\overline{d}_{\mathcal{P}}$, the maximal distance to the path $\max(d_{\mathcal{P}})$, and the average velocity ($\overline{v}$) of the robot for the different controllers and tracks.

\begin{table}
	\footnotesize
	\caption{Comparing path tracking controllers. Best/worst values are highlighted in green/red. State-of-the-art controllers are separated from controllers implemented for this work by an additional horizontal line.}
	\label{tab:performance}
	\centering
	\setlength{\fboxsep}{1pt}
\begin{tabular}{llccc}

	\toprule
	Controller                         & Track                  & $\overline{d}_{\mathcal{P}}$ in $[m]$ & $\max(d_{\mathcal{P}})$ in $m$   & $\overline{v}$ in $[ms^-1]$      \\
	\midrule
	\multirow[c]{2}{*}{$C_{MPPI}$}     & $\mathcal{P}_{cos}$    & 0.083                                 & 0.242                            & 0.561                            \\
	                                   & $\mathcal{P}_{square}$ & 0.173                                 & 0.481                            & 0.512                            \\
	\cline{1-5}
	\multirow[c]{2}{*}{$C_{RPP}$}      & $\mathcal{P}_{cos}$    & 0.110                                 & 0.337                            & \colorbox{green!70!black}{0.995} \\
	                                   & $\mathcal{P}_{square}$ & 0.271                                 & 0.734                            & \colorbox{green!70!black}{0.949} \\
	\cline{1-5}
	\multirow[c]{2}{*}{$C_{DWB}$}      & $\mathcal{P}_{cos}$    & \colorbox{green!70!black}{0.021}      & \colorbox{green!70!black}{0.059} & \colorbox{red!70}{0.209}         \\
	                                   & $\mathcal{P}_{square}$ & \colorbox{green!70!black}{0.038}      & \colorbox{green!70!black}{0.221} & \colorbox{red!70}{0.188}         \\
	\cline{1-5}
	\midrule
	\multirow[c]{2}{*}{$C_{PP}$}       & $\mathcal{P}_{cos}$    & 0.067                                 & 0.202                            & 0.498                            \\
	                                   & $\mathcal{P}_{square}$ & 0.164                                 & 0.539                            & 0.490                            \\
	\cline{1-5}
	\multirow[c]{2}{*}{$C_{RL}$}       & $\mathcal{P}_{cos}$    & 0.059                                 & 0.227                            & 0.544                            \\
	                                   & $\mathcal{P}_{square}$ & 0.305                                 & 1.375                            & 0.730                            \\
	\cline{1-5}
	\multirow[c]{2}{*}{$\hat{C}_{RL}$} & $\mathcal{P}_{cos}$    & \colorbox{red!70}{0.838}              & \colorbox{red!70}{2.875}         & 0.784                            \\
	                                   & $\mathcal{P}_{square}$ & \colorbox{red!70}{1.092}              & \colorbox{red!70}{2.753}         & 0.805                            \\
	\cline{1-5}
	\multirow[c]{2}{*}{$C_{S}$}        & $\mathcal{P}_{cos}$    & 0.058                                 & 0.192                            & 0.543                            \\
	                                   & $\mathcal{P}_{square}$ & 0.182                                 & 0.708                            & 0.492                            \\
	\cline{1-5}
	\multirow[c]{2}{*}{$\hat{C}_{S}$}  & $\mathcal{P}_{cos}$    & 0.165                                 & 0.589                            & 0.589                            \\
	                                   & $\mathcal{P}_{square}$ & 0.198                                 & 0.936                            & 0.500                            \\

	\bottomrule
\end{tabular}

	\vspace{-0.25cm}
\end{table}
% 1. Compare MPPI, RPP, DWB
% When comparing the \ac{mppi}, \ac{rpp}, and \ac{dwb} controllers, we observe that \ac{dwb} outperforms the other controllers in terms of tracking error and maximal distance to the path caused by the minimal velocity.
% Determining suitable configuration parameters for this controller was challenging, resulting in a suboptimal performance.
% However, it underlines our motivation to enable the use of high-performance controllers in safety-critical applications.
%
% The increased performance of those is exemplified by \ac{mppi} as well, which achieves the lowest tracking error and distance to the path at the cost of reduced velocity.
% The \ac{rpp} controller, on the other hand, shows a good compromise between tracking error and velocity.
\Cref{tab:performance} shows that \ac{dwb} has the lowest tracking error and lowest $\max(d_{\mathcal{P}})$ value, but its average velocity is limited.
This is caused by misconfiguration of the controller and underlines the challenge of finding suitable parameters.
\ac{mppi} achieves the lowest tracking error and distance to the path while maintaining acceptable velocity.
\ac{rpp}, however, balances tracking error and velocity best.

% 2. Compare PP to MPPI, RPP, DWB
% This supports our decision for using the Pure Pursuit algorithm as the basis for the high-assurance controller $C_{PP}(x)$, which achieves similar performance.
% Note that the velocity is constrained to $\SI{0.5}{\meter\per\sec}$ by design to ensure safety.
% This, however, results in a reduced maximal deviation to the path compared to \ac{mppi} and \ac{rpp}.
% ($C_{RL}(x)$ and $\hat{C}_{RL}(x)$).
The pure pursuit controller $C_{PP}$ achieves similar performance to \ac{mppi}, but with a constrained velocity of $\SI{0.5}{\meter\per\sec}$ to ensure safety.
Given its design, the results of \ac{rpp} indicate that better results could be achieved by $C_{PP}(x)$ as well.

% 3. Compare RL to PP, MPPI, RPP, DWB
% For the high-performance controller, we opted for an \ac{rl}-based controller ($C_{RL}(x)$ and $\hat{C}_{RL}(x)$).
% The prematurely terminated training of $\hat{C}_{RL}(x)$ results in a significant increase in the maximal distance to the path compared to $C_{RL}(x)$, which matches the observations in the simulation, \cmpr{fig:pytorch_failing}.
% The adaptiveness of the fully trained $C_{RL}(x)$ is reflected in the varying average velocity, which is increased for the squared track compared to the cosine track.
% On the other hand, its maximal distance to the path is higher than for the other controllers, which underlines the necessity for the Simplex architecture once again.
The \ac{rl}-based high-performance controllers $C_{RL}(x)$ and $\hat{C}_{RL}(x)$ show the necessity for the Simplex architecture.
The prematurely terminated training of $\hat{C}_{RL}(x)$ results in a significantly increased value of $\max(d_{\mathcal{P}})$ compared to $C_{RL}(x)$, which matches the observations in the simulation, \cmpr{fig:pytorch_failing}.
On the other hand, the fully trained $C_{RL}(x)$ shows an increased average velocity for the squared track compared to the cosine track and thereby underlines the potential of complex controllers.
Nevertheless, even $C_{RL}(x)$ has a value of $\max(d_{\mathcal{P}}) > \SI{1.0}{\meter}$ for the squared track, which underlines the necessity for the Simplex architecture.

% 5.1 Compare Simplex to others
% Thus, for both versions($C_S(x)$ and $\hat{C}_S(x)$) of the Simplex controller, we observe that the maximal distance to the path is reduced; always maintaining the safety requirements.
% $C_S(x)$ achieves the minimal average tracking error (excluding \ac{dwb}) for $\mathcal{P}_{cos}$ while having a similar velocity to \ac{mppi}.
% Furthermore, we can observe that the average velocity of $\SI{0.544}{\meter\per\sec}$ matches the one of $C_{RL}(x)$, meaning that the decision module did not need to switch to the high-assurance controller.
% Vice versa, for $\mathcal{P}_{square}$, $C_S(x)$ obtained an average velocity of $\SI{0.490}{\meter\per\sec} \leq \SI{0.5}{\meter\per\sec}$, which shows that the decision module needed to switch to the high-assurance controller to maintain the safety requirement.
% This is caused by the sharp turns in the track, which cause the relative angle between robot and path to exceed the threshold.
% Such a situation is exemplary shown in \cref{fig:simplex_assurance}.
The Simplex controllers ($C_S(x)$ and $\hat{C}_S(x)$) reduce $\max(d_{\mathcal{P}})$ to less than $\SI{0.9}{\meter}$ as planned and thereby maintain the safety requirement.
$C_S(x)$ achieves the minimal average tracking error (excluding \ac{dwb}) for $\mathcal{P}_{cos}$ while having a similar velocity to \ac{mppi}.
Furthermore, we can observe that the average velocity of $\SI{0.543}{\meter\per\sec}$ (almost) matches the one of $C_{RL}(x)$, meaning that the decision module did not need to switch to the high-assurance controller.
In contrast, for $\mathcal{P}_{square}$, the decision module switches to the high-assurance controller to maintain safety, as shown in \cref{fig:simplex_assurance}, which underlines the correct operation of the Simplex architecture once again.
However, one needs to note that the $\max(d_{\mathcal{P}})$ value of $\SI{0.936}{\meter}$ for $\hat{C}_S(x)$ on the squared track is above $\SI{0.681}{\meter}$, which was given by the conservative approximation of $\mathcal{X}_{S|C_{PP}}$, \cmpr{subsec:decision_module}.
This is caused by the sharp turns in the track, creating worst-case scenarios not covered by the reachability analysis.
While this does not violate the safety requirement, it underlines the necessity for in-depth analysis of the reachability set to guarantee safety.

The results, however, show limitations of the Simplex architecture as well.
While it enables the safe deployment of (suboptimal) \ac{rl} controllers, the performance, when measured as the average velocity, is reduced.
One reason for this is the conservative approximation of $\mathcal{X}'_{S|C_{PP}}$, in turn caused by the conservative convergence criteria of the reachability analysis.
A greater convex hull would allow for keeping the high-performance controller active for a longer time and thereby increase the overall performance of the Simplex controller.
This is also caused by the employed pure pursuit algorithm, whose simplicity enables safety guarantees but limits its \ac{roa} and in turn the set $\mathcal{X}'_{S|C_{PP}}$ as well.
Furthermore, it causes an increased dwell time of $T_d = \SI{12.45}{\sec}$, causing the Simplex controller to remain in high-assurance mode for an extended period of time.
Using average dwell time~\mycite{xiang2022necessary} could reduce switching intervals and enhance high-performance controller utilization, but demands stricter monitoring to ensure safe transitions from high-assurance to high-performance mode.

We finalized our evaluation by deploying the Simplex controller $\hat{C}_S$ on the Clearpath Husky robot in a real-world scenario.
A recording is available at \url{https://youtu.be/Fvf_wx9YJYo}.
%\textcolor{red}{A recording is submitted along with the paper and will be made available at YouTube, we will share the link with the camera-ready version of the article.}

%%%%
%% Conclusions
%%%%
\section{Conclusions}
\label{sec:conclusions}

We presented the need for adaptivity in path tracking controllers for increasing their performance and highlighted the gap of deploying them safely.
While safety and stability guarantees are provided by traditional controllers, these properties are challenged by increased complexity and, recently, \ac{rl}-based controllers.
To bridge this gap, we built on the Simplex architecture~\mycite{sha2001using} to propose a path tracking controller that combines a high-performance controller with a high-assurance controller to ensure safety at all times.

With that, our contribution is twofold.
Firstly, we described a design procedure for Simplex-based controllers, including a discussion on deriving safe switching strategies for the decision module based on the \ac{roa} of the high-assurance controller and the overall safety requirements.
Secondly, we applied the procedure to a differential drive robot's path tracking problem and evaluated the controller in simulation before real-world implementation.

Despite showing safety, restrictive switching conditions caused by conservative approximations of the high-assurance controller's \ac{roa} and similarly conservative usage of dwell time reduced the performance of the Simplex controller and highlight the architecture's limitation:
The performance gains possible have to be within the limits of the high-assurance controller's ability to ensure safety.

\section{Future Work}
\label{sec:future_work}

The highlighted limitations of the Simplex controller suggest several directions for future work.
For improving performance, less conservative approaches to designing the decision module and its switching conditions have to be researched.
This includes more accurate approximations of the high-assurance controller's \ac{roa} but also evaluating the use of average dwell time to enable the Simplex controller to switch more frequently.

Beyond that, architectural improvements could be made by using not only a single high-assurance controller but a set of controllers that can be switched between based on the current situation.
This can be used to not only separate concerns between performance and safety but also between environmental and contextual situations, effectively creating a more versatile and adaptable controller.

% \section*{Appendix}
% The following abbreviations are used in this manuscript:
\begin{acronym}[ECU]

	\acro{ann}[ANN]{Artificial Neural Network}
	\acro{airgemm}[AIRGEMM]{AI and Robotics for GeoEnvironmental Modeling and Monitoring}
	\acro{ai}[AI]{Artificial Intelligence}
	\acro{app}[APP]{Adaptive Pure Pursuit}
	\acro{ar}[AR]{Augmented Reality}
	\acro{agv}[AGV]{Autonomous Ground Vehicle}
	\acro{asv}[ASV]{Autonomous Surface Vehicle}
	\acro{bc}[BC]{Behavioral Cloning}
	\acro{cave}[CAVE]{Cave Automatic Virtual Environment}
	\acro{clf}[CLF]{Common Lyapunov Function}
	\acro{cbf}[CBF]{Control Barrier Function}
	\acro{dwa}[DWA]{Dynamic Window Approach}
	\acro{dwb}[DWB]{Dynamic Window Approach Version B}
	\acro{gps}[GPS]{Global Positioning System}
	\acro{gnss}[GNSS]{Global Navigation Satellite System}
	\acro{lidar}[LiDAR]{Light Detection and Ranging}
	\acro{ml}[ML]{Machine Learning}
	\acro{mppi}[MPPI]{Model Predictive Path Integral}
	\acro{mpc}[MPC]{Model Predictive Control}
	\acro{nn}[NN]{Neural Network}

	\acro{odd}[ODD]{Operational Design Domain}
	\acro{radar}[Radar]{Radio Detection and Ranging}
	\acro{rpp}[RPP]{Regulated Pure Pursuit}
	\acro{pp}[PP]{Pure Pursuit}

	\acro{hmd}[HMD]{Head-Mounted Display}
	\acro{hmi}[HMI]{Human Machine Interface}

	\acro{rl}[RL]{Reinforcement Learning}
	\acro{roa}[RoA]{Region of Attraction}
	\acro{ros}[ROS]{Robot Operating System}
	\acro{ros2}[ROS2]{Robot Operating System Version 2}
	\acro{sonar}[Sonar]{Sound Navigation and Ranging}

	\acro{tcp}[TCP]{Transmission Control Protocol}
	\acro{teb}[TEB]{Timed Elastic Band}
	\acro{td3}[TD3]{Twin Delayed Deep Deterministic Policy Gradient}
	\acro{td3bc}[TD3+BC]{Twin Delayed Deep Deterministic Policy Gradient with Behavioral Cloning}
	% \acro{sonar}[sonar]{sound navigation and ranging}
	\acro{tubaf}[TUBAF]{Freiberg University of Mining and Technology}

	\acro{vr}[VR]{Virtual Reality}
	\acro{xr}[XR]{Extended Reality}
\end{acronym}

%\section*{References}
%\printbibliography
%\balance

\bibliography{bibliography.bib}

\end{document}